# Minnorm training: an algorithm for training over-parameterized deep neural networks

**Yamini Bansal**
Harvard University
ybansal@g.harvard.edu

**Madhu Advani**
Harvard University
madvani@fas.harvard.edu

**David D Cox**
Harvard University
MIT-IBM Watson AI Lab
davidcox@fas.harvard.edu
david.d.cox@ibm.com

**Andrew M Saxe**
Harvard University
asaxe@fas.harvard.edu

## Abstract

In this work, we propose a new training method for finding minimum weight norm solutions in over-parameterized neural networks (NNs). This method seeks to improve training speed and generalization performance by framing NN training as a constrained optimization problem wherein the sum of the norm of the weights in each layer of the network is minimized, under the constraint of exactly fitting training data. It draws inspiration from support vector machines (SVMs), which are able to generalize well, despite often having an infinite number of free parameters in their primal form, and from recent theoretical generalization bounds on NNs which suggest that lower norm solutions generalize better. To solve this constrained optimization problem, our method employs Lagrange multipliers that act as integrators of error over training and identify 'support vector'-like examples. The method can be implemented as a wrapper around gradient based methods and uses standard back-propagation of gradients from the NN for both regression and classification versions of the algorithm. We provide theoretical justifications for the effectiveness of this algorithm in comparison to early stopping and $L_2$-regularization using simple, analytically tractable settings. In particular, we show faster convergence to the max-margin hyperplane in a shallow network (compared to vanilla gradient descent); faster convergence to the minimum-norm solution in a linear chain (compared to $L_2$-regularization); and initialization-independent generalization performance in a deep linear network. Finally, using the MNIST dataset, we demonstrate that this algorithm can boost test accuracy and identify difficult examples in real-world datasets.

## 1 Introduction

Deep neural networks are often over-parameterized relative to their training dataset size [17, 18, 25, 14, 30, 9, 1, 23]. Nevertheless, they still generalize well to novel inputs. This generalization ability has been traced to two interrelated aspects of the optimization problem: first, a large body of work has shown that the complexity of the function instantiated by a neural network is related to the norm of its weights [22, 4, 5, 2] suggesting that networks will generalize well to the extent that the optimization process ends with reasonably small weights. Second, recent results on the dynamics of gradient descent learning in the high dimensional regime have explained why gradient descent dynamics naturally yield low norm solutions in the underdetermined regime, even with infinite training time, provided that networks are initialized with small weights [1]. Together, these findings provide one explanation for why vanilla deep networks often generalize well without any sort of regularization (beyond starting from small initializations).

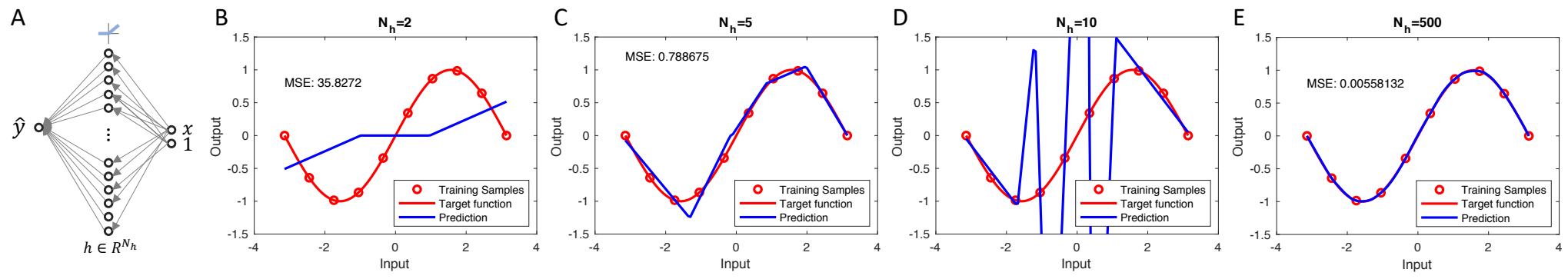

Figure 1: Over-parameterization and generalization via minimum norm solutions. (A) A simple ReLU network with random first layer weights and trained second layer weights. The network receives a scalar input and bias term, and is trained to minimize squared error on ten points (red circles) from a target sinusoid (red curve) shown in panels B-E. (B-E) Blue curves show example functions learned by networks with differing numbers of hidden units $N_h = \{2, 5, 10, 500\}$. Networks in panels B, C, & D show the standard progression from underfitting the training data to overfitting, with a happy medium in panel C. However the large network in panel E generalizes best. This network has $50\times$ more parameters than training examples but generalizes well, because among the infinity of solutions attaining zero training error we have chosen a low norm solution. In this work we derive training algorithms for nonlinear deep networks that explicitly seek minimum norm solutions in the underdetermined, accurate label regime, allowing good generalization in large networks.

As a concrete example, Fig 1 depicts a simple ReLU network receiving a scalar input (panel A), trained from ten samples of a sinusoidal target function. As the number of hidden units is increased from 2 to 10, the network shows the usual progression from underfitting to overfitting (panels B-D). However, a large network with 500 hidden units generalizes best in this simple scenario, because among the many weight configurations that attain zero training error, we have selected a low norm solution.

These prior results have focused on explaining the success of vanilla deep networks trained using gradient descent, the nearly ubiquitous workhorse of deep learning. In this work, we present a novel training algorithm specifically designed to seek low norm solutions for the underdetermined, accurate-label regime in which many deep networks operate. In particular, deep NNs are typically trained to minimize an unconstrained objective, for instance, the cross-entropy error on a corpus of training data. Here we investigate an alternative paradigm of constrained optimization in over-parametrized networks, wherein training proceeds by requiring a neural network to minimize the size of its weights subject to the constraint that it exactly fits its training data. This scheme, which brings Lagrangian constrained optimization methods into deep learning, behaves dramatically differently from standard gradient descent, showing no apparent overtraining, improved generalization performance compared to standard SGD training, insensitivity to weight initialization, and improvements in norm-based generalization bound values.

In Section 2, we introduce the main constrained optimization formulation and discuss its connection to support vector machines. As it is based on constrained optimization, it is applicable in the low label noise, underdetermined regime in which a deep network can exactly fit the training dataset. However, this is typically the case in deep learning applications. In Section 3, we introduce four illustrative toy problems to demonstrate the operation of the method. First, we show that the method successfully identifies support vector-like training examples that control the separating hyperplane. Next, we show that the formulation differs from $L_2$-regularization by yielding fast convergence to minimum norm solutions without shifting the location of training loss minima. Third, we analyze the method when applied to shallow networks, and contrast it with the standard binary cross entropy loss. We show that the method can give much faster convergence to the max-margin hyperplane. Finally, we analyze how, in the underdetermined regime, the Minnorm training procedure decays weights in the 'frozen subspace' in which no data lies, yielding improved generalization and reduced sensitivity to weight initialization compared to standard SGD training. In Section 4 we apply the method to the MNIST dataset, where it trains in similar time to SGD but achieves better test accuracy and exhibits no overtraining.

## 2 Overview of Minnorm training

The method we describe here is conceptually similar to the optimization problem underlying the SVM algorithm, which yields robust classification by finding a maximum margin solution [6]. This robustness allows SVMs to generalize well when trained on finite datasets despite being effectively over-parameterized. For instance [10] describes a kernel for which SVM learning is equivalent to finding a max margin classifier for an infinite width shallow ReLU neural network with random first layer weights. Despite having an infinite-dimensional feature space, fitting this model does not lead to high generalization error because, among all decision boundaries which correctly classify the training samples, the SVM optimization algorithm selects the one with maximum margin. Unfortunately, despite their benefits, SVMs do not scale well with the number of training samples when solved in the dual form. In this paper we propose a method for taking advantage of the robust performance of maximum margin-style solutions while maintaining the training speed and scalability of deep NNs.

The essence of the method is to formulate an optimization problem which, expressed informally, corresponds roughly to "minimize the complexity of the network's input-output map subject to fitting all training points exactly." Because the complexity of a network's input-output map is complicated, we replace this with a surrogate: the norm of the weights in each layer. From this high level formulation, it is also apparent that such an approach only applies when (a) the network can fit all training points exactly, i.e., it is underdetermined; and (b) the labels are accurate, such that exactly fitting them is desirable. If the labels themselves are noisy, then the traditional unconstrained optimization formulations might be preferable to the methods we develop here. We now describe several instantiations of this high level formulation.

First we note that the standard hard margin SVM is an example of this approach. Consider training a linear SVM to learn a set of weights $W \in \mathbb{R}^{1\times N}$ which provide the max-margin classification of the input-output pairs $(x^\mu \in R^{N\times 1}, y^\mu \in \{-1, 1\})$ for a dataset containing $P$ examples $\mu = 1, ..., P$ with $\hat{y}^\mu = Wx^\mu$. The maximum margin optimization problem is easily converted to the equivalent problem of finding the smallest-norm weights subject to the constraint that all examples are bounded by a constant margin:

$$\min ||W||_2^2 \quad \text{s.t.} \quad \hat{y}^\mu y^\mu \geq 1 \quad \text{for} \quad \mu = 1, ..., P. \tag{1}$$

This primal form of the SVM is convex.

Here we propose an extension of this optimization to deep non-linear networks by instead minimizing the sum of the squared norm of weights $W_l$ in each layer, $l = 1, \cdots, D$

$$\min \sum_l ||W_l||_F^2 \quad \text{s.t.} \quad \hat{y}^\mu = y^\mu \quad \text{for} \quad \mu = 1, ..., P. \tag{2}$$

The above equations are appropriate for the regression context where $y^\mu \in \mathbb{R}$ take continuous scalar values, but we will generalize the method to handle classification shortly. Note that the optimization is no longer convex since $\hat{y}(x)$ is a depth $D$ NN. In the context of neural network training, this optimization is based on the idea that NNs with small weights implement simpler functions, and hence a minimum norm solution is likely to generalize well. The fact that lower norm NNs implement simpler functions is formalized by a variety of norm-based generalization bounds for deep NNs [21], [22], [4], and we note that it depends on using activation functions which are simpler near the origin. Critically, Minnorm training differs from simply adding an $L_2$ penalty to bias the network towards low weight solutions, because such a penalty will move the fixed points of the objective function and increase training error.

To solve (2), we introduce Lagrange multipliers $\alpha^\mu$ for each example and form the Lagrangian:

$$\mathcal{L}_\rho(W, \alpha) = \sum_{l=1}^{D} \frac{1}{2}||W_l||_F^2 + \sum_{\mu=1}^{P} \alpha^\mu (y^\mu - \hat{y}^\mu) + \frac{\rho}{2}\|\hat{y} - y\|^2. \tag{3}$$

The Minnorm optimization problem corresponds to maximizing the Lagrangian over $\alpha$ while simultaneously minimizing over weights. Optimizing the first two terms in the above Lagrangian directly ($\rho = 0$) can work in practice, but it leads to oscillations as we discuss in Section 3.2. To damp these we use an augmented Lagrangian method (see e.g. [7]) by including a quadratic term modulated by scalar parameter $\rho$ to improve convergence. This additional term contributes nothing to the gradient

**Algorithm 1:** Minnorm training: Regression

**Data:** $\{x^\mu, y^\mu\}_{\mu=1}^P$
**Initialize:** $\alpha^\mu = 0 \quad \mu = 1, ..., P$; $W$ = your favorite initialization
**Parameters**: $\rho$, $\eta$, $s$, $Q$
**for** *number of epochs* **do**
  **for** *number of mini-batches* **do**
    $\mathcal{S}$ = set of minibatch indices of size $Q$
    $\mathcal{L}_\rho^Q(W,\alpha) = \sum_{l=1}^D \frac{1}{2}||W_l||_F^2 + \sum_{\mu\in\mathcal{S}} \alpha^\mu(y^\mu - \hat{y}^\mu) + \frac{\rho}{2}\sum_{\mu\in\mathcal{S}}(y^\mu - \hat{y}^\mu)^2$
    $W^{t+1} = W^t - \eta\partial_W \mathcal{L}_\rho^Q(W^t, \alpha)$
    $\alpha_{t+1}^\mu = \alpha_t^\mu + s(y^\mu - \hat{y}^\mu(W^{t+1})) \quad \mu \in \mathcal{S}$;

of the Lagrangian at the Minnorm solution, assuming that all the constraints are satisfied (training predictions match data), and hence it does not impact the fixed points of the optimization process.

We perform a pair of iterative updates similar to the dual-ascent algorithm to optimize for the network weights. First

$$W^{t+1} = W^t - \eta\partial_W \mathcal{L}_\rho(W^t, \alpha), \tag{4}$$

where $\eta$ is the step size for weight updates. The above equation corresponds to gradient descent on the Lagrangian and can be applied to arbitrary deep networks via standard automatic differentiation. This weight update step can be replaced by multiple gradient steps or alternative neural network training approaches such as momentum [28] or Adam [15]. The second step is the update of the Lagrange multipliers with step size $s$:

$$\alpha_{t+1}^\mu = \alpha_t^\mu + s(y^\mu - \hat{y}^\mu(W^{t+1})). \tag{5}$$

These updates change the cost function being optimized in the first update, and hence differ from prior methods which have brought SVM training into deep networks by employing the hinge loss with standard unconstrained optimization [29]. From the updates of the $\alpha^\mu$, they can be loosely interpreted as the integrated errors made by the network for each example over the course of training. Pseudocode is given in Alg. 1. The framework we just described is designed for regression, but can easily be extended to classification as we now discuss.

To derive algorithms for the binary and multi-class classification problems, we keep the same weight norm penalty but adjust the requirements of the constraints (algorithmic details are given in Appendix A). In the binary setting, the most important change is that the equalities in the optimization formulation become inequalities, which in turn leads to a constraint that the Lagrange multipliers are non-negative. Therefore we replace the gradient update of the $\alpha^\mu$ with projected gradient descent to enforce this non-negativity. In the multi-class setting we extend the binary algorithm using a one-vs-all classification scheme which introduces Lagrangian parameters for each sample and output channel (indexed as $\alpha_i^\mu$). Pseudocode for this multi-class formulation is given in Alg. 2. Extensions such as allowing for slack variables are discussed in Appendix C, and have been found effective in a slightly different setting in [29] which replaces the final softmax layer of a neural network with an SVM classifier.

## 3 Analysis of Minnorm training in simplified settings

The Minnorm optimization scheme introduces new parameters and hyperparameters into the optimization process. In this section we provide interpretations and analyses of these parameters by investigating simple tractable settings. First, we show that the Lagrange multipliers can be interpreted similarly to support vector weights in standard SVM training. Hence only non-zero Lagrange multiplier variables influence the location of the decision boundary, and intuitively identify the 'hardest' examples in the dataset. Second, we investigate the convergence properties of the algorithm in a minimally deep setting, and compare it to the standard practice of adding an $L_2$ weight decay penalty to the typical unconstrained optimization formulation. We show that in the $L_2$ penalty case, the location of the training error optimum is shifted in inverse proportion to the speed of convergence to the minimum norm solution, whereas our Minnorm algorithm achieves fast convergence without moving the location of the minimum. Third, we analyze the speed of convergence to the max-margin

**Algorithm 2:** Minnorm training: Multi-class classification

**Data:** $\{x^\mu, y^\mu\}_{\mu=1}^P$
Initialize: $\alpha^\mu = 0 \quad \mu = 1, ..., P, i = 1, ..., N_o$, $W$ = your favorite initialization;
Parameters: $\rho$, $\eta$, $s$, $Q$
**for** *number of epochs* **do**
  **for** *number of minibatches* **do**
    $\mathcal{S}$ = set of minibatch indices of size $Q$
    $\mathcal{L}_\rho^Q(W, \alpha) =$
    $\sum_{l=1}^D \frac{1}{2}||W_l||_F^2 + \sum_{\mu\in\mathcal{S}} \sum_{i=1}^{N_o} \alpha^\mu(1 - y_i^\mu \hat{y}_i^\mu) + \frac{\rho}{2} \sum_{\mu\in\mathcal{S}:\alpha^\mu>0} \sum_{i=1}^{N_o}(1 - y_i^\mu \hat{y}_i^\mu)^2$
    $W^{t+1} = W^t - \eta \partial_W \mathcal{L}_\rho^Q(W^t, \alpha)$
    $\alpha_{i,t+1}^\mu = \pi^+ \left[\alpha_{i,t}^\mu + s(1 - y_i^\mu \hat{y}_i^\mu(W^{t+1}))\right] \quad \mu \in \mathcal{S}; \quad i = 1, ..., N_o,$
    where $\pi^+$ is a projection onto the positive orthant.

hyperplane in a simple setting with shallow networks under the standard binary cross entropy (BCE) loss, and under Minnorm training. We find that Minnorm training can converge substantially faster than gradient descent on the cross entropy loss. Finally, it has been shown that gradient-based training of underdetermined deep networks can yield a frozen subspace in which weights do not change because no data lies in these directions [1]. In this subspace, the initial random weights remain indefinitely and can harm generalization performance. Using deep linear neural networks in a student-teacher formalism, we show that Minnorm training successfully decays weights in these frozen directions, yielding generalization that matches explicit pruning of the frozen subspace; and no overtraining.

### 3.1 Fixed point structure and support vector-like variables

The major difference between Minnorm training and standard unconstrained optimization formulations is that optimization occurs with an augmented set of Lagrange multiplier variables that enforce the zero training error constraint. Here we seek greater understanding of these quantities by investigating the structure of the fixed points if training converges.

The fixed points of the dynamics in Eqns. (4)-(5) are

$$W_l = \sum_{\mu=1}^P \alpha^\mu \frac{\partial \hat{y}^\mu}{\partial W_l} \quad \text{for} \quad l = 1, \cdots, D, \tag{6}$$

$$y^\mu = \hat{y}^\mu \quad \text{for} \quad \mu = 1, \cdots, P. \tag{7}$$

The latter equation arises from the Lagrange multipliers and shows that if the optimization converges, it attains zero training error. The first expression yields insight into the weights at the optimal solution. In particular, it states that the optimal minimum norm weights are a linear combination of the matrices $\frac{\partial \hat{y}^\mu}{\partial W_l}$. The matrix $\frac{\partial \hat{y}^\mu}{\partial W_l}$ is the derivative of the network's output on example $\mu$ with respect to matrix $W_l$. The weightings of the linear combination are the Lagrange multipliers $\alpha^\mu$. Even without knowing the precise values of the $\alpha^\mu$, these equations reveal substantial structure in the ultimate solutions. The crucial fact is that each weight matrix $W_l$ is a combination of just $\tilde{P}$ terms, where $\tilde{P}$ is the number of nonzero $\alpha^\mu$; but it lives in a $N_{l+1} \times N_l$ space, where $N_l$ is the number of neurons in layer $l$. Hence if $\tilde{P} < N_{l+1} \times N_l$, there will be a subspace of the weights which is zeroed out in the minimum norm solution. Moreover, we note that the same $\alpha^\mu$ apply to different layers. This means that the total vectorized weights Vec$(W_1, \cdots, W_D)$ are a linear combination of the $\tilde{P}$ vectors Vec $\left(\frac{\partial \hat{y}^\mu}{\partial W_1}, \cdots, \frac{\partial \hat{y}^\mu}{\partial W_D}\right)$. Hence there will be a zeroed subspace of the weights provided that $\tilde{P} < N_{tot}$, where $N_{tot}$ is the total number of parameters in the network. This gestures to the potential reduction in number of effective parameters in deep networks when trained in this way.

Finally, we note that as the only contributions to weights come from nonzero $\alpha^\mu$, they play a role analogous to support vector coefficients in standard SVM formulations (and coincide exactly in the shallow case). To demonstrate this point, Fig. 2 compares support vectors identified in a binary classification task using standard SVM training with nonzero Lagrange multipliers found in deep nonlinear networks with Minnorm training: both methods find the same support vectors. Also, the deep multilayer nonlinear networks still settle on a simple, nearly linear classification boundary, suggesting that their complexity is not causing over-fitting. The connection of the Minnorm Lagrange multipliers to support vectors has several potential applications. First, non-support vectors need not be backpropagated, as their contribution to the weight update is zero. This offers a potential run time speed up for datasets with a low number of important examples, as is often seen in practice. Second, these variables are related to influence functions describing the importance of each example to a particular classification [16]. Third, the examples that are support vectors can be illuminating, identifying the hardest (or possibly mislabeled) training examples, as we show later in numerical experiments.

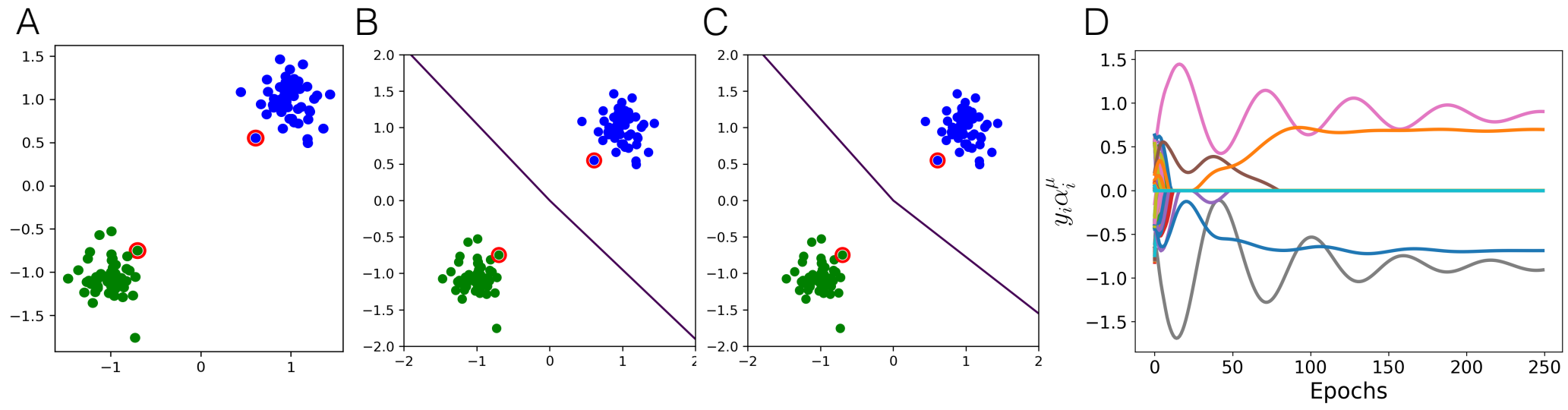

Figure 2: Support vector-like quantities in nonlinear deep networks. A standard linear SVM and several deep networks are trained on a simple linearly separable binary classification. (A) SVM with linear kernel. Identified support vectors highlighted in red. (B) Minnorm training of 1 hidden layer, 100 hidden unit ReLU network. Examples with nonzero $\alpha^\mu$ at the end of training highlighted in red. (C) Minnorm training of 3 hidden layer, 100 hidden units/layer network. Examples with nonzero $\alpha^\mu$ at the end of training highlighted in red. (D) Lagrange multipliers $y_i\alpha_i^\mu$ (as in Eq. 42) during training. The Lagrange multiplier variables in Minnorm training serve a similar role to support vectors, identifying those training samples that set the boundary between classes. These can be a small fraction of the total dataset, as shown in this case.

### 3.2 Training dynamics in a simple linear chain

Moving beyond the fixed points, we now turn to convergence dynamics. In particular we contrast Minnorm training with the common practice of adding an $L_2$ penalty on the weights and solving an unconstrained optimization problem. We examine the tractable case of a simple toy problem consisting of learning a linear chain of three neurons. This model serves as a minimal example of depth. We present this network with a single example with scalar input and label $(x, y)$, which can be fit perfectly by a linear chain that can learn functions of the form $\hat{y} = w_2 w_1 x$. However, the loss surface for this network is non-convex [3, 12, 24] and the solution manifold for this simple case is the hyperbola $w_1 w_2 = 1$. The minimum norm solution in this case is where the weights at each layer are equal ($w_1 = w_2 = \pm 1$).

In Fig 3, we compare the weight trajectories during training for vanilla gradient descent, $L_2$-regularization and the Minnorm algorithm. We now theoretically analyze the dynamics of gradient descent with $L_2$ regularization and compare it to the dynamics of Minnorm training to better understand their convergence rates, accuracy, and stability.

#### 3.2.1 $L_2$ regularized gradient descent

In the scalar two neuron chain with a single example, the traditional unconstrained training loss function with an $L_2$ penalty (weight decay) takes the form:

$$l(w_1, w_2) = \frac{1}{2}\left(y - w_2 w_1 x\right)^2 + \frac{\gamma}{2}\left(w_2^2 + w_1^2\right), \tag{8}$$

and the gradient descent dynamics in the continuous time limit have the form:

$$\tau \dot{w}_1 = (y - w_2 w_1 x) w_2 x - \gamma w_1, \qquad \tau \dot{w}_2 = (y - w_2 w_1 x) w_1 x - \gamma w_2, \tag{9}$$

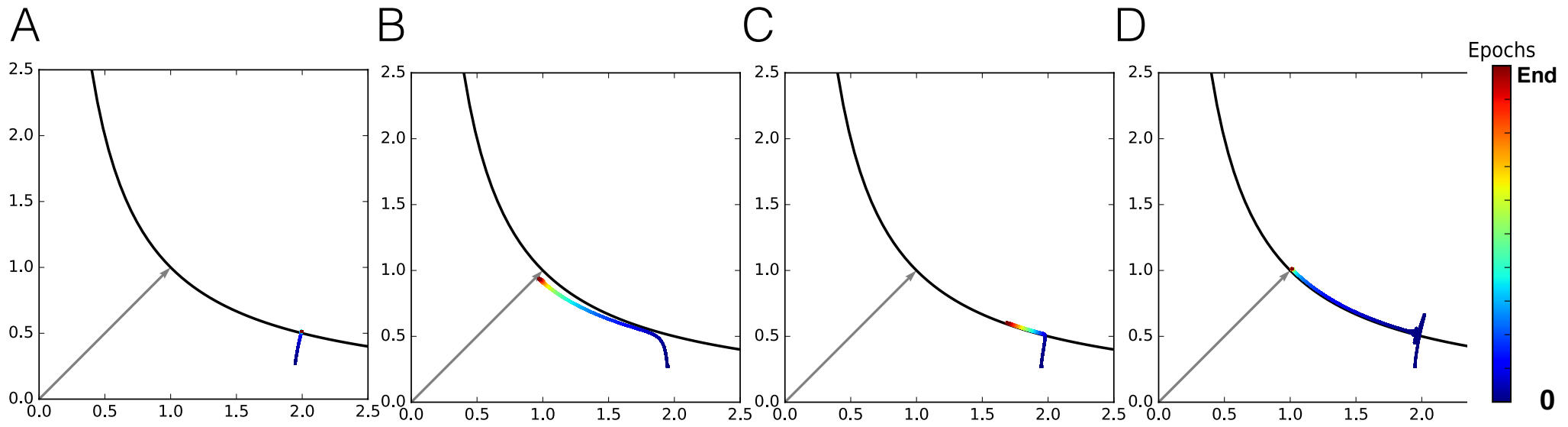

Figure 3: Comparison of the weight trajectories during vanilla gradient descent, $L_2$-regularization and the Minnorm training for a toy model starting from the same initial point. (A) SGD (5K epochs) converges quickly to the zero training error manifold but will not approach a low norm solution with balanced weights across layers. (B) $L_2$-regularization ($\gamma = 0.05$, 20K epochs) converges fast but to a point off of the zero training error manifold (C) $L_2$-regularization ($\gamma = 0.005$, 20K epochs) reaches the solution manifold but is progressing towards the minimum norm solution glacially. (D) Minnorm (5K epochs, $\rho = s = 100$, $\eta = 0.01$) avoids this speed-accuracy tradeoff and needs only 5K epochs to converge to the minimum norm solution on the zero training error manifold.

where $\tau = \frac{1}{\eta}$ is the inverse learning rate. If weights start imbalanced ($w_2(0) \neq w_1(0)$), this must be corrected to achieve the minimum norm solution. To track the speed at which they achieve balance, we define the weight gap (between layers) as $\Delta = \frac{1}{2}(w_2^2 - w_1^2)$, so that

$$\frac{d\Delta}{dt} = \frac{d}{dt}\frac{1}{2}(w_2^2 - w_1^2) = w_2\dot{w}_2 - w_1\dot{w}_1 = -\frac{2\gamma}{\tau}\frac{1}{2}(w_2^2 - w_1^2) = -\frac{2\gamma}{\tau}\Delta. \tag{10}$$

Thus, the weight gap decays exponentially with a timescale of $\frac{\tau}{2\gamma}$ so that regularization strength is inversely proportional to convergence time. However, regularization also shifts the fixed points of the dynamics which satisfy:

$$(y - w_2 w_1 x) = \frac{\gamma w_1}{w_2 x}, \qquad (y - w_2 w_1 x) = \frac{\gamma w_2}{w_1 x}. \tag{11}$$

Setting the RHS of the preceding two equations equal to each other implies that for $\gamma > 0$, at the fixed point we have $w_2 = w_1$ which implies that the fixed point satisfies $y - \hat{y} = \frac{\gamma}{x}$, thus the convergence time-scale scales inversely with the training error due to regularization. This trade-off can be observed in Fig 3 B and C, where low regularization causes slow learning close to the solution manifold and large regularization causes faster learning which shifts off the solution manifold.

### 3.2.2 Minnorm dynamics

We now analyze the dynamics of the Minnorm algorithm, via dual ascent on (3) in the two neuron chain, which yields the continuous time dynamics:

$$\tau\dot{\alpha} = r(y - \hat{y}), \tag{12}$$

$$\tau\dot{w}_1 = -w_1 + \alpha w_2 - \rho(\hat{y} - y)w_2 x, \tag{13}$$

$$\tau\dot{w}_2 = -w_2 + \alpha w_1 - \rho(\hat{y} - y)w_1 x. \tag{14}$$

Note that we define $r = \tau s = \frac{s}{\eta}$. We can use (13) and (14) to again show that the weight gap will decay exponentially but this time with a time-scale of $\tau/2$ independent of $\rho, r$:

$$\frac{d\Delta}{dt} = \frac{d}{dt}\frac{1}{2}(w_2^2 - w_1^2) = w_2\dot{w}_2 - w_1\dot{w}_1 = \frac{1}{\tau}\left(-w_2^2 + w_1^2\right) = -\frac{2\Delta}{\tau}. \tag{15}$$

This also implies that the "balanced" condition of $w_2 \approx w_1 = w$ is a reasonable simplification and we will use the notation $\bar{w} = w_2 w_1 = w^2$. We can use the fact that $\frac{d}{dt}w^2 = 2w\dot{w}$ to show that (13) and (14) imply:

$$\tau\dot{\bar{w}} = -2\bar{w} + 2\alpha\bar{w} + 2\rho(y - \hat{y})\bar{w}. \tag{16}$$

In the simple case of $y = 1, x = 1$, our dynamics simplify to the pair of coupled differential equations:

$$\tau\dot{\bar{w}} = 2\bar{w}\left(\alpha - 1 + \rho(1-\bar{w})\right), \qquad \tau\dot{\alpha} = r(1-\bar{w}). \tag{17}$$

The fixed point of the dynamics occurs at $(w = 1, \alpha = 1)$ and we linearize the dynamics about this fixed point (see Appendix B for the full calculation and, e.g., [27] section 6.3 for more detail on this method). This analysis leads to the eigenvalues:

$$\lambda = \frac{1}{2\tau}\left(-2\rho \pm \sqrt{4\rho^2 - 8r}\right). \tag{18}$$

The fixed point will be stable when the real part of both eigenvalues is negative, which occurs for any $\rho$ and $r$ greater than zero, and will be a stable spiral point when the eigenvalues have an imaginary component which corresponds to $r \geq \frac{1}{2}\rho^2$. Thus, a stable spiral fixed point will occur under the choice of large $r$ and small $\rho$.

In the special case of $\rho = 0$, the eigenvalues become purely imaginary $\lambda = \pm\frac{i\sqrt{2r}}{\tau}$ implying that the fixed point is a neutrally stable center and that the solutions will oscillate at a rate proportional to $\sqrt{r}$. In this case we can analytically analyze the dynamics by taking the ratio of the two dynamical equations and collecting like terms on each side of the equality. We then integrate to show:

$$\ln(\bar{w}) - \bar{w} + c_0 = \frac{2}{r}\left(\frac{\alpha^2}{2} - \alpha\right). \tag{19}$$

See Appendix B for derivation details and note that $c_0$ is a constant determined by the initial values of the weights and Lagrange multipliers. The above equations suggest that the oscillations will be more pronounced in weight space if $r$ is small. The practical lessons emerging from this analysis are to keep updates of the Lagrange multipliers relatively fast (large $r$ and therefore large $s$) and add some amount of $\rho$ for stabilization.

### 3.3 Convergence speed to the max-margin hyperplane in shallow networks

A key motivation behind our method is the goal of finding a max-margin classification boundary similar to that of an SVM. Here we study the speed of convergence to the max-margin solution in the case of a shallow (zero hidden layer) neural network. Traditional networks trained with the cross-entropy loss converge to the maximum margin solution, but this convergence is extremely slow, roughly $\log(t)$ [26, 20, 8]. By contrast, we illustrate through a simple example that the Minnorm training algorithm makes this 'implicit bias' of regular training explicit in its formulation, which allows it fast exponential convergence, roughly $O(e^{-\frac{t}{\tau}})$. We consider the simple case of a binary classification in two dimensions with one example per class. Without loss of generality, we take the positive class input to be $x^1 = \frac{1}{\sqrt{2}}(1,1)$ and the negative class input to be $x^2 = \frac{1}{\sqrt{2}}(-1,-1)$ as shown in Fig. 4A. We now analyze the convergence speed in both cases.

#### 3.3.1 Convergence speed of gradient descent on the binary cross-entropy loss

We first study the dynamics of learning in the traditional setting of cross-entropy error, which for this setting reduces to logistic regression. The convergence rate of gradient descent to the max margin hyperplane is known [26] to be $O(\log(t))$. Here we furnish explicit solutions in this simple setting.

We consider a shallow neural network with weight vector $W = [w_1 w_2]$ and sigmoidal output $\hat{y}^\mu = \sigma(Wx^\mu)$ where $\sigma(u) = \frac{1}{1+e^{-u}}$. The binary cross entropy loss is

$$\mathcal{L}_{BCE} = \sum_{\mu=1}^{P} -\tilde{y}^\mu \log \hat{y}^\mu - (1-\tilde{y}^\mu)\log(1-\hat{y}^\mu)$$

where $\tilde{y}^\mu = y^\mu/2 + 1/2$ is the the output label $\{-1, 1\}$ recoded to $\{0, 1\}$. Differentiating with respect to the weights yields the gradient descent update:

$$W_{t+1} = W_t + \eta \sum_{\mu=1}^{P} \left[\tilde{y}^\mu - \sigma(Wx^\mu)\right] {x^\mu}^T.$$

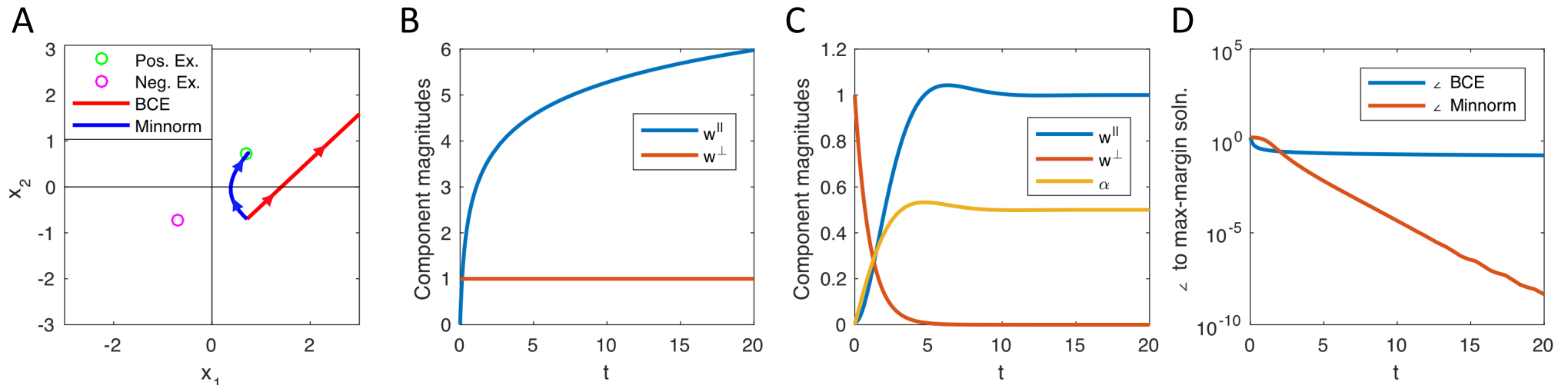

Figure 4: Convergence to the max-margin solution in shallow networks. (A) A binary classification task in two dimensions with one positive and one negative example. Trajectories of the network's weight vector over learning starting from $(1/\sqrt{2}, -1/\sqrt{2})$ are shown for a shallow network learning with the binary cross entropy loss (red), and for the Minnorm method (blue). (B) The binary cross entropy network grows weights in the direction parallel to the max-margin vector, but leaves components of the weights perpendicular to the max-margin unchanged. (C) Minnorm training rapidly converges to the max-margin weight vector. Weights parallel to the max-margin vector converge exponentially to their fixed point, with some ringing. Weights perpendicular to the max-margin decay exponentially. (D) As the BCE network's weights grow, the angle to the max-margin hyperplane decreases slowly at rate $O(1/\log(t))$. The Minnorm weights converge exponentially fast to the max-margin solution (note log-scale y-axis). Simulation parameters: For the BCE network, $\tau = 1/10$. For the Minnorm network, $\tau = 1, r = 1/4$.

Taking the continuous time limit (valid for small $\eta$), for our dataset we have

$$\begin{aligned} \tau \dot{W} &= \left[1 - \sigma(Wx^1)\right] {x^1}^T + \left[0 - \sigma(Wx^2)\right] {x^2}^T && (20) \\ &= \left[\frac{e^{Wx^1}}{1 + e^{Wx^1}}\right] {x^1}^T + \sigma(-Wx^1){x^1}^T && (21) \\ &= 2\sigma(-Wx^1){x^1}^T, && (22) \end{aligned}$$

where we have used the fact that $x^1 = -x^2$. We now decompose the weights as $W(t) = w^{||}\frac{1}{\sqrt{2}}(1,1)^T + w^{\perp}\frac{1}{\sqrt{2}}(1,-1)^T$ into a component parallel and perpendicular to the direction of the maximum margin hyperplane. We have

$$\tau \dot{W} = \dot{w}^{||}\frac{1}{\sqrt{2}}(1,1)^T + \dot{w}^{\perp}\frac{1}{\sqrt{2}}(1,-1)^T = 2\sigma(-w^{||})\frac{1}{\sqrt{2}}(1,1)^T,$$

and hence

$$\begin{aligned} \tau \dot{w}^{||} &= 2\sigma(-w^{||}), && (23) \\ \tau \dot{w}^{\perp} &= 0. && (24) \end{aligned}$$

The component in the direction of the max margin hyperplane will therefore continue to grow without bound ($w^{||} \to \infty$), while the component orthogonal to the max margin hyperplane $w^{\perp} = w^{\perp}(0)$ remains unchanged. Eqn. (23) is separable and may be integrated to find the time required to go from some initial value $w^{||}(0) = w_0^{||}$ to the value $w^{||}(t) = w_f^{||}$:

$$\begin{aligned} \tau \frac{dw^{||}}{dt} &= \frac{2}{1 + e^{w^{||}}} && (25) \\ \int dt &= \int_{w_0^{||}}^{w_f^{||}} \frac{\tau}{2}\left(1 + e^{w^{||}}\right) dw^{||} && (26) \\ t &= \frac{\tau}{2}\left[w_f^{||} - w_0^{||} + e^{w_f^{||}} - e^{w_0^{||}}\right] && (27) \end{aligned}$$

From this it is easy to see that at long times when $w^{||}$ is large, it will grow logarithmically with time, $w^{||}(t) = O(\log t)$, consistent with the findings of [26]. The angle between $w(t)$ and the max-margin

solution $\hat{w} = x^1$ thus scales as

$$\sin\theta = \frac{w^{\perp}}{\sqrt{(w^{||})^2 + (w^{\perp})^2}} = \frac{1}{\frac{w^{||}}{w^{\perp}}\sqrt{(1+\frac{w^{\perp}}{w^{||}})^2}} = O\left(\frac{1}{\log(t)}\right). \tag{28}$$

It follows that in the large $t$ limit, the angle between $w(t)$ and the max-margin classifier will scale as $\theta = O\left(\frac{1}{\log(t)}\right)$. These dynamics are shown in Fig. 4B and D.

#### 3.3.2 Convergence speed of Minnorm training

In this setting, the continuous time dynamics have the form:

$$\tau\dot{W} = -W + \sum_{\mu} \alpha^{\mu} y^{\mu} x^{\mu} = -W + (\alpha^1 + \alpha^2)x^1, \tag{29}$$

and both Lagrange multipliers have the form

$$\dot{\alpha}^{\mu} = r(1 - w^{||}x^1) \quad \text{s.t.} \quad \alpha^{\mu} \geq 0. \tag{30}$$

By decomposing the weight dynamics we see:

$$\tau\dot{w}^{||} = -w^{||} + \alpha^1 + \alpha^2 \tag{31}$$

$$\tau\dot{w}^{\perp} = -w^{\perp} \tag{32}$$

$$\tau\dot{\alpha}^{\mu} = r(1 - w^{||}) \quad \text{s.t.} \quad \alpha^{\mu} \geq 0 \quad \mu = 1, 2. \tag{33}$$

This is a system of linear differential equations. From the perpendicular dynamics, we see that this component of the weights is decoupled from the other variables and decays exponentially with time scale $\tau$. The parallel dynamics remain coupled to the Lagrange multiplier dynamics. We note that the dynamics for all $\alpha^{\mu}$ are the same, and hence if they are initialized identically (here to zero), they will remain equal and we have $\alpha = \alpha^1 = \alpha^2$. Therefore we have the two variable system

$$\tau\dot{w}^{||} = -w^{||} + 2\alpha \tag{34}$$

$$\tau\dot{\alpha} = r(1 - w^{||}) \quad \text{s.t.} \quad \alpha \geq 0. \tag{35}$$

with a fixed point at $(\alpha = \frac{1}{2}, w^{||} = 1)$. For small perturbations around this fixed point we may neglect the inequality constraint on $\alpha$, and the dynamics correspond to eigenvalues $\lambda = \frac{1}{2\tau}\left(-1 \pm \sqrt{1 - 8r}\right)$. In the practically relevant regime where $r > 0$ and $\tau$ finite, both eigenvalues have negative real part and the continuous time system is always stable. If $8r > 1$, the dynamics oscillate as a stable spiral with exponential decay as $O(e^{-\frac{t}{2\tau}})$. Supposing the parameters are chosen to critically damp the system ($r = \frac{1}{8}$) we see that the parallel component will converge with time scale $2\tau$ to its fixed point, $w^{||} = 1 + O\left(e^{-\frac{t}{2\tau}}\right)$. Combining this with the form of $w^{\perp}$ implies that the angle of the solution to the max margin converges as $\theta = O\left(e^{-\frac{t}{\tau}}\right)$ at large times. These dynamics are shown in Fig. 4B and D. In sum, while traditional training techniques can eventually attain minimal norm or max margin solutions, the Minnorm algorithm based on dual ascent can be substantially faster.

### 3.4 Pruning the frozen subspace in deep linear networks

As a final example, we consider generalization dynamics using the linear student-teacher framework studied in [1]. In this toy model, a teacher network with randomly drawn Gaussian weights $\bar{W} \sim \mathcal{N}(0, I)$ generates a dataset by labeling a set of $P$ randomly drawn *i.i.d.* $N$-dimensional Gaussian inputs $x^{\mu} \sim \mathcal{N}(0, \frac{1}{N}I)$. The scalar output of the teacher network, plus noise $\epsilon \sim \mathcal{N}(0, 1/\text{SNR})$, yields the target values $y^{\mu} = \bar{W}x^{\mu} + \epsilon$ that are fed to the student network. The student network is trained with a mean-squared error loss. These assumptions on the training data and teacher parameters in the student-teacher setting allow for exact calculation of the test error, that is, the expected $L_2$ error of the student on a new example. The analysis in [1] shows that, in under-determined shallow networks, there is a frozen subspace of the weights that do not change under vanilla gradient descent. The weight initialization in this subspace persists indefinitely, harming generalization performance. Thus in under-determined networks, generalization error scales with the size of initial weight scale.

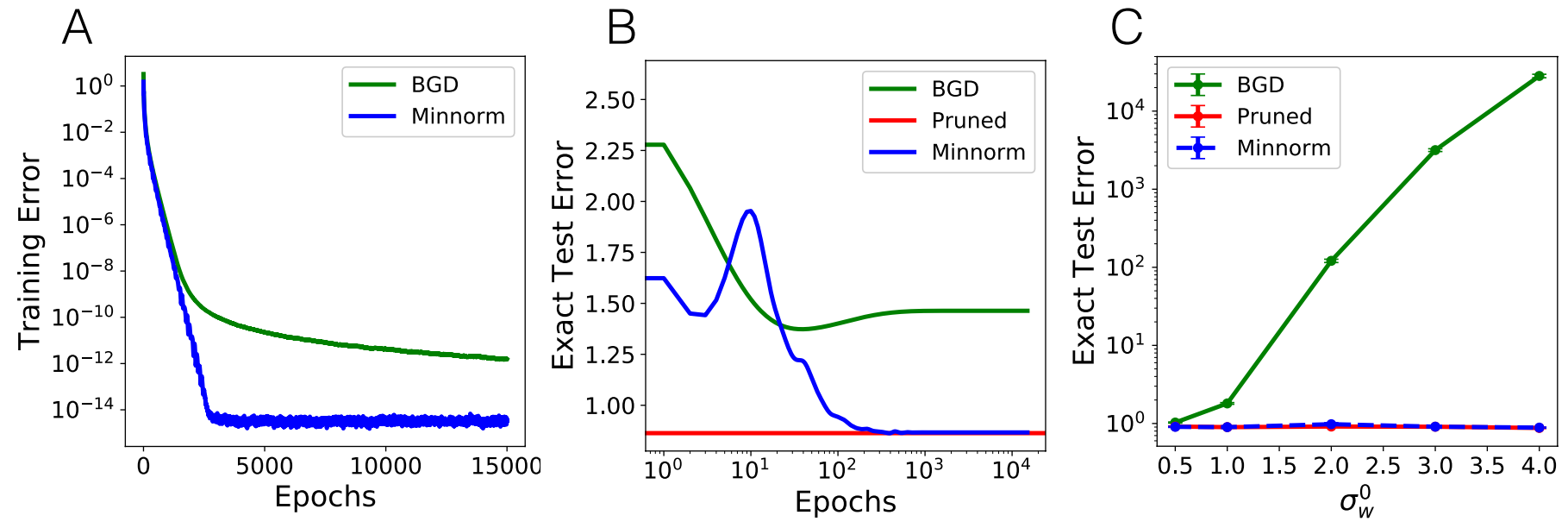

Figure 5: Minnorm training in underdetermined deep linear neural networks. Three layer deep linear networks with 300 hidden units and 150 examples are trained on the student-teacher task used in [1] with and SNR (variance of teacher parameters divided by variance in noise) of 5. (A) Training error dynamics for Minnorm and BGD training for weight initialization scale $\sigma_w^0 = 1$. Minnorm training makes exponentially rapid progress. (B) Test error dynamics. The Minnorm algorithm attains better final performance than BGD, eventually reaching the performance of a network where weights in the frozen subspace have been pruned. In contrast to BGD, the Minnorm dynamics show no overtraining. (C) Test error as a function of initial weight scale $\sigma_w^0$, averaged over 20 runs. In contrast to vanilla BGD, Minnorm training removes the dependence of test error on initialization weight scale. We note that, for fast training speeds, a weight scale of $\sigma_w^0 = 1$ is the current best practice for linear networks [13, 24], but this incurs a cost in test error when used with BGD.

Here we show that Minnorm training decays weights in these directions, removing the dependence of generalization error on initialization in this setting.

Figure 5 shows training and test dynamics for vanilla batch gradient descent and Minnorm training. As shown in panels A and B, Minnorm dynamics train quickly, yield better generalization performance and show no overtraining in this overparameterized case. The Minnorm algorithm eventually attains the performance of the linear model in which we have hand-pruned all weights in the frozen subspace orthogonal to the input examples, demonstrating that the method successfully decays weights in these irrelevant directions. Finally, panel C shows the asymptotic test error as a function of the standard deviation of initial weights, showing that unlike BGD, Minnorm training achieves the same test error regardless of initialization.

To analyze this effect theoretically, consider the case where the student is an $D$ layer deep linear network that produces output $\hat{y} = \prod_{l=1}^{D} W_l x$, where $W_L$ is a row vector because we have restricted to a single output. From the fixed points of Minnorm training in Eqn. (6), and using the fact that $\frac{\partial \hat{y}^\mu}{\partial W_1} = (\prod_{l=2}^{D} W_l)^T x^{\mu^T}$ for the deep linear network, we have $W_1 = \sum_{\mu=1}^{P} \alpha^\mu (\prod_{l=2}^{D} W_l)^T x^{\mu^T}$. The overall input-output map of the network is thus

$$\prod_{l=1}^{D} W_l = \Big(\prod_{l=2}^{D} W_l\Big)\Big(\prod_{l=2}^{D} W_l\Big)^T \sum_{\mu=1}^{P} \alpha^\mu x^{\mu^T} = \Big\|\prod_{l=2}^{D} W_l\Big\|_2^2 \sum_{\mu:\alpha^\mu \neq 0} \alpha^\mu x^{\mu^T}. \tag{36}$$

Hence at convergence, the overall weights must lie in the $\tilde{P}$-dimensional subspace spanned by the inputs corresponding to non-zero Lagrange multipliers, and the subspace orthogonal to these inputs is completely zeroed out.

## 4 Training deep nonlinear networks on real-world tasks

In this section we apply the Minnorm algorithm to nonlinear deep networks in the multiclass setting using the MNIST dataset, comparing its performance to minibatch gradient descent and $L_2$ weight decay. We train a fully connected network with 2 hidden layers and 800 hidden units with ReLU activations on 50K examples for each algorithm. We then pick the optimal stopping time (and weight decay in the L2 regularized setting) on the validation set (10K examples), and report the test accuracy for the same stopping time (and weight decay) on the test set (10K examples). Details about the hyperparameters can be found in Appendix D. In Table 1 we report the validation error and test error averaged over 10 runs. The Minnorm algorithm with minibatch gradient descent performs best. Interestingly, Minnorm with full batch also provides an improvement over previous methods, suggesting that this method could be amenable to large batch training. Our experiments here focus on testing the merit of different optimization methods on a fixed vanilla network architecture, rather than attaining state of the art performance, and we note that the Minnorm training method may be applied with standard regularizers such as dropout, batchnorm, and more complicated architectures like convolutional networks [19, 25].

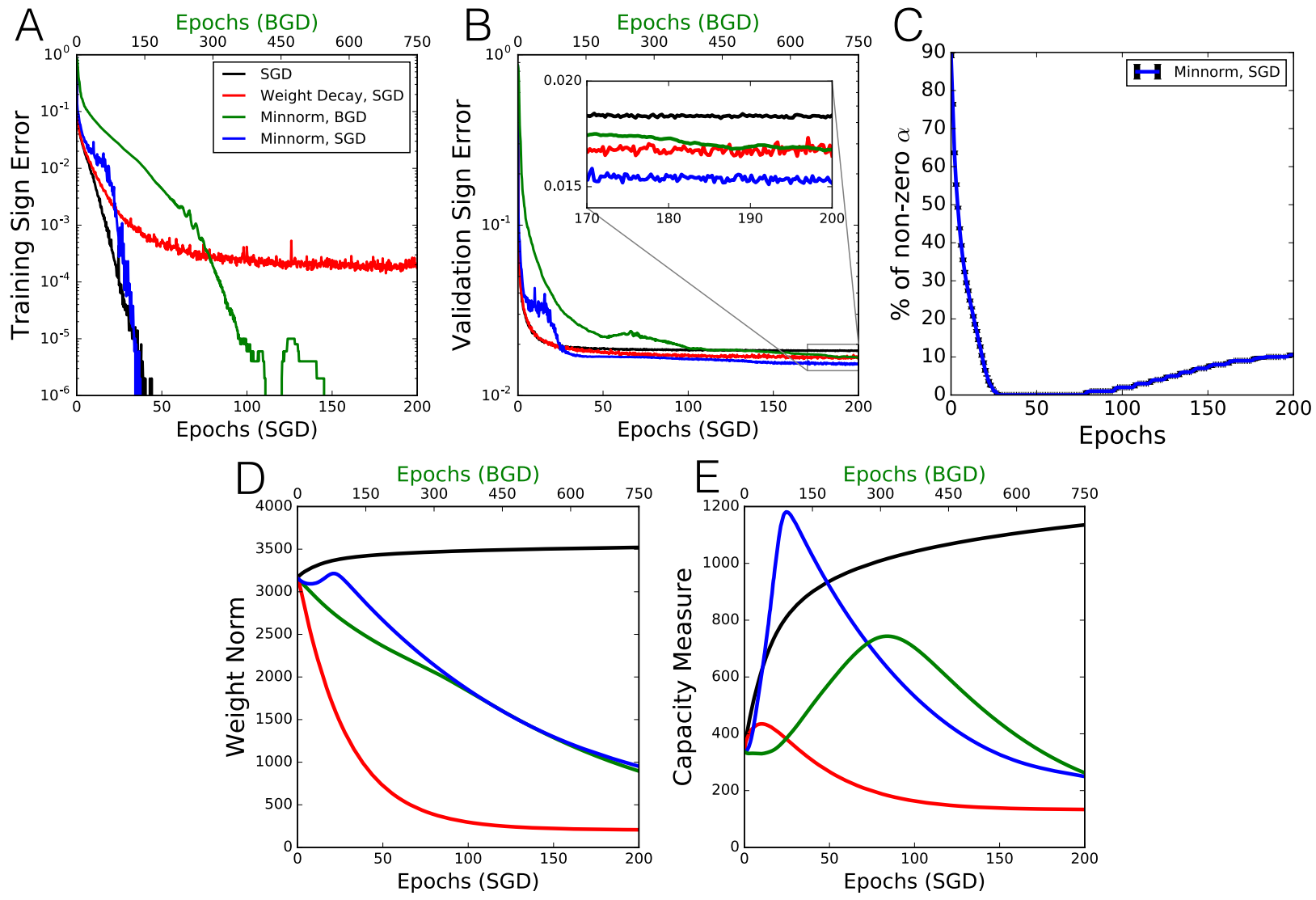

Figure 6: Results in nonlinear deep networks on MNIST. Networks contained 2 hidden layers of 800 ReLU units each, training hyperparameters (in particular, $L_2$ weight decay) optimized on a validation set. (A) Training error. (B) Test error. (C) Percentage of examples that have non-zero $\alpha_i$ for at least one class. About 90% of the dataset has zero $\alpha$ for all classes after 30 epochs of training. (D) Weight norms over training ($\sum_{l=1}^{3} \|W\|_F^2$). (E) Norm-based capacity bound of [22] ignoring constant factors. Minnorm training is about as fast as SGD but generalizes better, and attains a 5-fold improvement in generalization bound.

| | Val. Error (%) | Train Error | Test Error (%) |
|---|---|---|---|
| Vanilla SGD | 1.78 ± 0.057 | 0.0008 ± 0.0024 | 1.82 ± 0.056 |
| Weight Decay | 1.55 ± 0.042 | 0.0136 ± 0.005 | 1.74 ± 0.051 |
| Minnorm, BGD | 1.56 ± 0.03 | 0.0 ± 0.0 | 1.52 ± 0.054 |
| Minnorm, SGD | **1.42** ± **0.052** | 0.0 ± 0.0 | **1.46** ± **0.063** |

Table 1: Generalization performance of SGD, Weight decay and Minnorm on MNIST.

Figure 6 confirms several of the qualitative results seen on toy datasets, including perfect training accuracy, better validation accuracy than $L_2$ weight penalties or vanilla SGD, and no over-training. The Minnorm algorithm finds substantially smaller weight norm solutions that still attain zero training error.

Minnorm training can also help improve generalization bounds based on weight norms. In Fig. 6 E, we plot a generalization error bound based on weight norm and stable rank [22]. Ignoring constant factors which would remain constant in our comparison, for our networks this metric is

$$\sqrt{\prod_{l=1}^{3} \|W_l\|_2^2 \sum_{l=1}^{3} \frac{\|W_l\|_F^2}{\|W_l\|_2^2}}.$$

Minnorm training improves this generalization bound over vanilla SGD by a factor of about five. While such norm-based bounds are still often vacuous (but see [2], they have been found to predict the qualitative behavior of generalization error [4, 22]. We note that our networks trained with weight decay achieve a strong generalization bound, but this is a bound on the difference between training and test accuracy. Because weight decay has worse training error than Minnorm, this leads to slightly worse test performance. Finding non-vacuous generalization bounds for neural networks is an active area of research [2, 11] and our training algorithm can tighten these bounds by minimizing the quantities of interest.

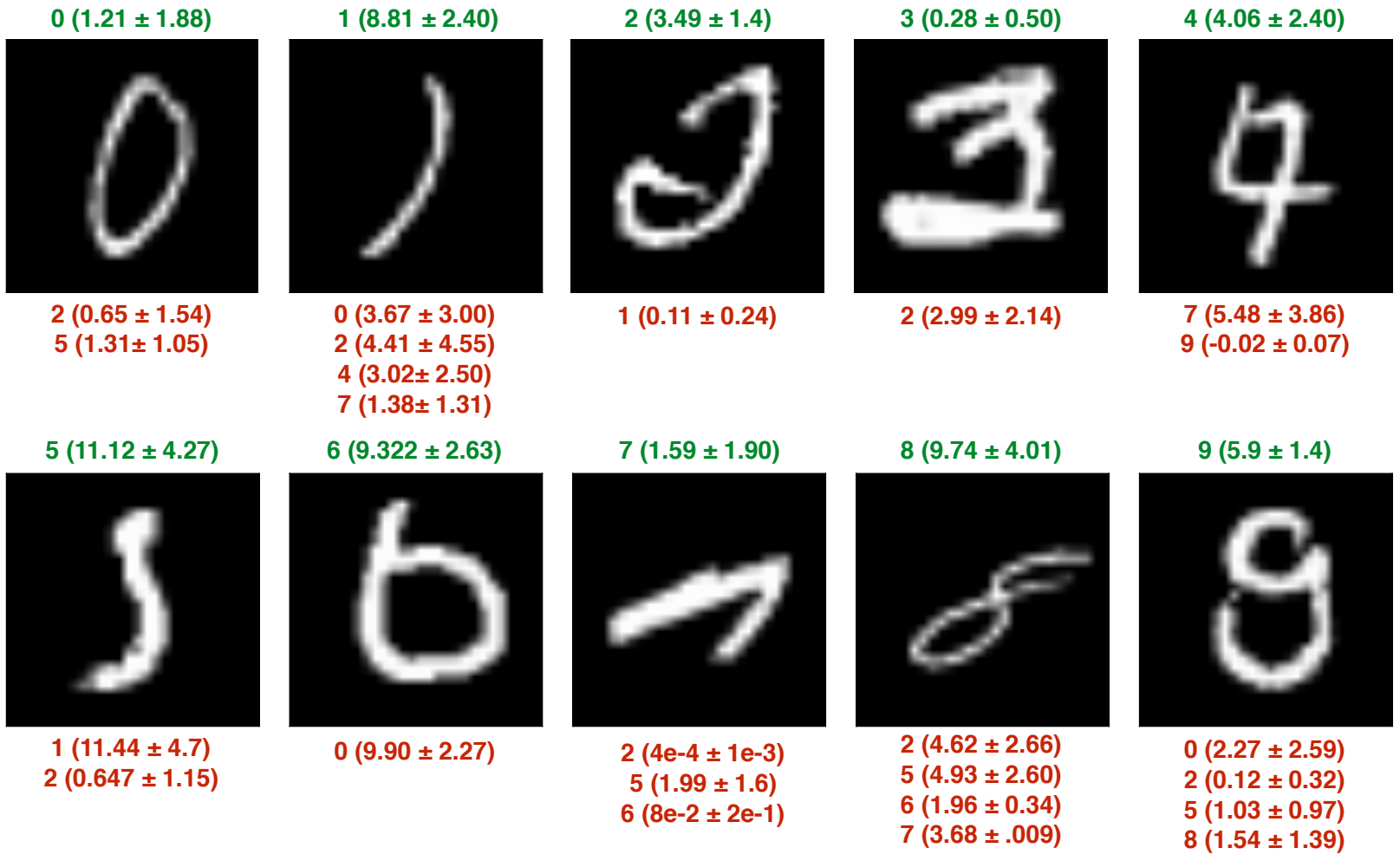

Figure 7: Generalized support vectors on the MNIST dataset. Examples for each digit class with non-zero $\alpha^\mu$ across 10 trials. The label and coefficient (in parentheses) for the correct class is in green above. Nonzero Lagrange multipliers for incorrect labels are in red below. Intuitively, support vectors are difficult examples that are easily confused with other categories.

Finally, in Fig. 7 we plot some examples which result in non-zero $\alpha_i^\mu$ at the end of training in all 10 trials. Since our formulation consists of 10 $\alpha_i^\mu$ (corresponding to the 10 digit classes) for each example $\mu$, a non-zero co-efficient of the incorrect class $\alpha_{i:\hat{y}_i \neq y}^\mu$ indicates the digit classes this $x^\mu$ maybe confused for. As we can see, these identified 'support vectors' do include ambiguous outliers.

## 5 Conclusion

In this work we have described a novel training procedure suitable for training large deep networks in the underdetermined, accurate label regime. Through theoretical analyses, we have shown several favorable properties of the algorithm, including fast convergence to minimum norm solutions without moving fixed points; identification of important examples through a support vector-like mechanism; and generalization benefits. The Minnorm algorithms introduced here hold promise in several additional directions which we have left as topics for future work. For one, the training procedure considered here can be combined with current regularizers such as dropout and batchnorm. In addition, the exact training error constraint can potentially be relaxed by including slack variables, which is particularly important for the regime of inaccurate/noisy labels. Finally, the zero support vectors could be removed from backpropagation, decreasing wall-clock training time. On MNIST, for instance, a substantial fraction of examples are not support vectors (Fig. 6 C). Given that large models operating in an accurate label regime are commonplace in current practice, we argue that our method presents a promising new direction for training neural network models.

### Acknowledgments

This work was supported by grant number IIS 1409097 from the National Science Foundation and IARPA contract D16PC00002. Andrew Saxe and Madhu Advani thank the Swartz Program in Theoretical Neuroscience at Harvard University.

## References

[1] M.S. Advani and A.M. Saxe. High-dimensional dynamics of generalization error in neural networks. *arXiv:1710.03667*, 2017.

[2] S. Arora, R. Ge, B. Neyshabur, and Y. Zhang. Stronger generalization bounds for deep nets via a compression approach. *arXiv:1802.05296*, 2018.

[3] P. Baldi and Y. Chauvin. Temporal Evolution of Generalization during Learning in Linear Networks. *Neural Computation*, 3:589–603, 1991.

[4] P.L. Bartlett, D.J. Foster, and M.J. Telgarsky. Spectrally-normalized margin bounds for neural networks. In *NIPS*, 2017.

[5] P.L. Bartlett and S. Mendelson. Rademacher and gaussian complexities: Risk bounds and structural results. *Journal of Machine Learning Research*, 3(Nov):463–482, 2002.

[6] Mikhail Belkin, Siyuan Ma, and Soumik Mandal. To understand deep learning we need to understand kernel learning. *arXiv:1802.01396*, 2018.

[7] S. Boyd, N. Parikh, E. Chu, B. Peleato, J. Eckstein, et al. Distributed optimization and statistical learning via the alternating direction method of multipliers. *Foundations and Trends in Machine learning*, 3(1):1–122, 2011.

[8] A. Brutzkus, A. Globerson, E. Malach, and S. Shalev-Shwartz. SGD Learns Over-parameterized Networks that Provably Generalize on Linearly Separable Data. *arXiv:1710.10174*, 2017.

[9] A. Canziani, A. Paszke, and E. Culurciello. An Analysis of Deep Neural Network Models for Practical Applications. *arXiv*, pages 1–7, 2017.

[10] Y. Cho and L.K. Saul. Kernel methods for deep learning. In *Advances in neural information processing systems*, pages 342–350, 2009.

[11] G.K. Dziugaite and D.M. Roy. Computing nonvacuous generalization bounds for deep (stochastic) neural networks with many more parameters than training data. *arXiv:1703.11008*, 2017.

[12] K. Fukumizu. Effect of Batch Learning In Multilayer Neural Networks. In *Proceedings of the 5th International Conference on Neural Information Processing*, pages 67–70, 1998.

[13] X. Glorot and Y. Bengio. Understanding the difficulty of training deep feedforward neural networks. *AISTATS*, 2010.

[14] K. He, X. Zhang, S. Ren, and J. Sun. Delving deep into rectifiers: Surpassing human-level performance on imagenet classification. *ICCV*, pages 1026–1034, 2016.

[15] D. Kingma and J. Ba. Adam: A Method for Stochastic Optimization. In *ICLR*, 2015.

[16] P.W. Koh and P. Liang. Understanding Black-box Predictions via Influence Functions. In *ICML*, 2017.

[17] S. Lawrence, C.L. Giles, and A.C. Tsoi. What Size Neural Network Gives Optimal Generalization? Convergence Properties of Backpropagation, 1996.

[18] Y. LeCun, Y. Bengio, and G. Hinton. Deep learning. *Nature*, 521:436–444, 2015.

[19] Y. LeCun, L. Bottou, Y. Bengio, and P. Haffner. Gradient-based learning applied to document recognition. *Proceedings of the IEEE*, 86(11):2278–2323, 1998.

[20] M.S. Nacson, J. Lee, S. Gunasekar, N. Srebro, and D. Soudry. Convergence of gradient descent on separable data. *arXiv:1803.01905*, 2018.

[21] B. Neyshabur, S. Bhojanapalli, D. Mcallester, and N. Srebro. Exploring generalization in deep learning. In *NIPS*. 2017.

[22] B. Neyshabur, S. Bhojanapalli, D. McAllester, and N. Srebro. A pac-bayesian approach to spectrally-normalized margin bounds for neural networks. *arXiv:1707.09564*, 2017.

[23] T. Poggio, K. Kawaguchi, Q. Liao, B. Miranda, L. Rosasco, X. Boix, J. Hidary, and H. Mhaskar. *CBMM Memo No 073*, 2017.

[24] A.M. Saxe, J.L. McClelland, and S. Ganguli. Exact solutions to the nonlinear dynamics of learning in deep linear neural networks. In Y. Bengio and Y. LeCun, editors, *ICLR*, Banff, Canada, 2014.

[25] K. Simonyan and A. Zisserman. Very Deep Convolutional Networks for Large-Scale Image Recognition. In *ICLR*, pages 1–14, 2015.

[26] D. Soudry, E. Hoffer, and N. Srebro. The implicit bias of gradient descent on separable data. *arXiv:1710.10345*, 2017.

[27] S.H. Strogatz. *Nonlinear dynamics and chaos: with applications to physics, biology, chemistry, and engineering*. CRC Press, 2018.

[28] I. Sutskever, J. Martens, G. Dahl, and G.E. Hinton. On the importance of initialization and momentum in deep learning. In *ICML*, 2013.

[29] Y. Tang. Deep Learning using Linear Support Vector Machines. In *ICML Workshop on Challenges in Representation Learning*, 2013.

[30] C. Zhang, S. Bengio, M. Hardt, B. Recht, and O. Vinyals. Understanding deep learning requires rethinking generalization. In *ICLR*, 2017.

# Appendix

## A Classification version of the algorithm

**Binary** Before considering multi-class problems, we begin with binary classification with data $y^\mu \in \{-1, 1\}$ where we define $f(x^\mu)$ to be the pre-activation in the final output neuron of a deep network. Then our Minnorm algorithm corresponds to the optimization:

$$\min \sum_l ||W_l||_F^2 \quad \text{s.t.} \quad y^\mu f(x^\mu) \geq 1 \quad \text{for} \quad \mu = 1, ..., P. \tag{37}$$

The Lagrangian takes the form:

$$L(W, \alpha) = \frac{1}{2} \sum_l ||W_l||_F^2 + \sum_\mu \alpha^\mu (1 - y^\mu f(x^\mu)). \tag{38}$$

In this setting of an inequality-constrained optimization we seek to find weights satisfying:

$$\hat{W} = \arg \min_W \max_{\alpha^\mu \geq 0} L(W, \alpha). \tag{39}$$

In order to deal with the positivity constraint on the Lagrange multipliers, we apply projected gradient descent (PGD), defining $\pi^+$ to be the projection onto the positive orthant. The PGD algorithm has the following two steps

$$W^{t+1} = W^t + \eta \sum_\mu \alpha_t^\mu y^\mu \partial_W f(x^\mu, W^t), \tag{40}$$

$$\alpha_{t+1}^\mu = \pi^+ \left[ \alpha_t^\mu + s(1 - f(x^\mu)) \right]. \tag{41}$$

**Multi-class** We can extend the binary classification algorithm to work on multi-class data. To do so consider output vector $y = \{-1, 1\}^{N_o}$ where one element index $i$ will correspond to the correct classifier: $y_i = 1$ and other indices $j \neq i$ satisfy $y_j = -1$. Note that this will lead to Lagrangian parameters for each sample and output channel (indexed as $\alpha_i^\mu$). This approach is typically known as a one-vs-all classification scheme in multiclass SVMs and is what is used for all multi-class results in this paper. The Lagrangian takes the form

$$L(W, \alpha) = \frac{1}{2} \sum_l \|W_l\|_2^2 + \sum_{\mu=1}^{P} \sum_{i=1}^{N_o} \alpha_i^\mu (1 - y_i^\mu f_i(x^\mu)) \tag{42}$$

where $N_o$ is the number of output neurons and $f_i(x)$ is the pre-activation of the $i^{th}$ output neuron. Other versions of multiclass classification in SVMs have straightforward analogues in our framework, and can require fewer Lagrange multiplier variables, but we leave this for future work. We now turn to an analysis of the dynamics of learning in simplified settings using this scheme.

# B Simple chain dynamics and stability analysis derivations

We can now analyze the dynamics of the Minnorm algorithm, via dual ascent on the augmented Lagrangian:

$$\begin{aligned} \tau\dot{\alpha} &= r(y-\hat{y}), & (43)\\ \tau\dot{w}_1 &= -w_1+\alpha w_2-\rho(\hat{y}-y)w_2 x, & (44)\\ \tau\dot{w}_2 &= -w_2+\alpha w_1-\rho(\hat{y}-y)w_1 x. & (45)\end{aligned}$$

Note that we are defining $r = s\tau = \frac{s}{\eta}$. We can use (44) and (45) to again show that the weight gap $\Delta$ will decay exponentially:

$$\frac{d\Delta}{dt} = \frac{d}{dt}\frac{1}{2}(w_2^2 - w_1^2) = w_2\dot{w}_2 - w_1\dot{w}_1 = \frac{1}{\tau}\left(-w_2^2 + w_1^2\right) = -\frac{2\Delta}{\tau}. \tag{46}$$

Thus time scale of the decay in weight gap is simply the inverse step size on the weights $\tau = \frac{1}{\eta}$. This also implies that the "balanced" condition of $w_2 \approx w_1 = w$ is a reasonable simplification and we will use the notation $\bar{w} = w_2 w_1 = w^2$. We can use the fact that $\frac{d}{dt}w^2 = 2w\dot{w}$ to show that (44) and (45) imply:

$$\tau\dot{\bar{w}} = -2\bar{w} + 2\alpha\bar{w} + 2\rho(y-\hat{y})\bar{w}. \tag{47}$$

In the simple case of $y = 1, x = 1$, our dynamics simplify to the pair of coupled differential equations:

$$\tau\dot{\bar{w}} = 2\bar{w}\left(\alpha - 1 + \rho(1-\bar{w})\right), \qquad \tau\dot{\alpha} = r(1-\bar{w}). \tag{48}$$

The fixed point of the dynamics occurs at $(w = 1, \alpha = 1)$. We can test the stability of this fixed point by adding a small perturbation of $\alpha = 1 + \delta\alpha$ and $\bar{w} = 1 + \delta\bar{w}$, so that

$$\delta\dot{\bar{w}} = \dot{\bar{w}} = \frac{2}{\tau}\left(\delta\alpha - \rho\delta\bar{w}\right), \qquad \delta\dot{\alpha} = \dot{\alpha} = -\frac{r}{\tau}\delta\bar{w}. \tag{49}$$

Thus, if we let $x = [\delta\bar{w}, \delta\alpha]^T$, we can write linearized dynamics as:

$$\dot{x} = \frac{1}{\tau}Tx, \quad \text{where} \quad T = \begin{pmatrix} -2\rho & 2 \\ -r & 0 \end{pmatrix}. \tag{50}$$

Diagonalizing $T$ yields the following constraint on the eigenvalues of T:

$$\lambda\tau(\lambda\tau + 2\rho) + 2r = 0. \tag{51}$$

Using the quadratic formula, the eigenvalues take the form:

$$\lambda = \frac{1}{2\tau}\left(-2\rho \pm \sqrt{4\rho^2 - 8r}\right). \tag{52}$$

Thus the fixed point will be stable for and $\rho$ and $r$ greater than zero, and will be a stable spiral point if $r \geq \frac{1}{2}\rho^2$. This suggests that a stable spiral fixed point will likely occur under the choice of large $r$ and small $\rho$.

If we consider the special case of $\rho = 0$, the eigenvalues become purely imaginary $\lambda = \pm\frac{i\sqrt{2r}}{\tau}$ indicating that the fixed point is a neutrally stable center and that the solutions will oscillate at a rate proportional to $\sqrt{r}$. In this case we can analytically analyze the dynamics by taking the ratio of the two dynamical equations:

$$\frac{d\bar{w}}{d\alpha} = \frac{1}{r}\frac{2\bar{w}(\alpha-1)}{1-\bar{w}}. \tag{53}$$

We can re-arrange this equation to collect like terms on each side of the equality:

$$\frac{d\bar{w}(1-\bar{w})}{\bar{w}} = \frac{2}{r}d\alpha(\alpha - 1). \tag{54}$$

We then integrate both sides of the expression which yields:

$$\ln(\bar{w}) - \bar{w} + c_0 = \frac{2}{r}\left(\frac{\alpha^2}{2} - \alpha\right), \tag{55}$$

where $c_0$ is a constant determined by the initial values of weights and Lagrange multipliers.

# C Alternative implementations

Here we include some alternative implementations and extensions of the Minnorm algorithm for future study.

## C.1 Adding slack terms

$$\min_{W,C} \quad \sum_l ||W_l||_F^2 + C\sum_\mu \xi^\mu \tag{56}$$

$$\text{s.t.} \qquad y^\mu f(x^\mu) \geq 1 - \xi^\mu, \quad \mu = 1, \cdots, P. \tag{57}$$

In this case, we can write the complete loss function without lagrange multipliers as:

$$\min_{W,C} \sum_l ||W_l||_F^2 + C\sum_{\mu=1}^{P} \max\left[1 - f(x^\mu)y_k^\mu, 0\right]. \tag{58}$$

## C.2 Soft-max implementation

In this setting we only have a single Lagrange multiplier to store for each example, regardless of the output dimension, but the algorithm involves some additional complexity in computing a max operation over output neurons. Here we simply enforce the condition that the correct classification is a certain amount larger than the next largest classification, thus if we let $i_\mu^*$ be the index of $y^\mu$ equal to one (corresponding to the correct class), then we require:

$$\hat{y}_{i_\mu^*} - \max_{j \neq i_\mu^*} \hat{y}_j \geq 1. \tag{59}$$

The Lagrangian in the soft-max setting will have the form:

$$L(W, \alpha) = \sum_l \|W_l\|_2^2 + \sum_{\mu=1}^{P} \alpha^\mu (1 - \hat{y}_{i_\mu^*} + \max_{j \neq i_\mu^*} \hat{y}_j). \tag{60}$$

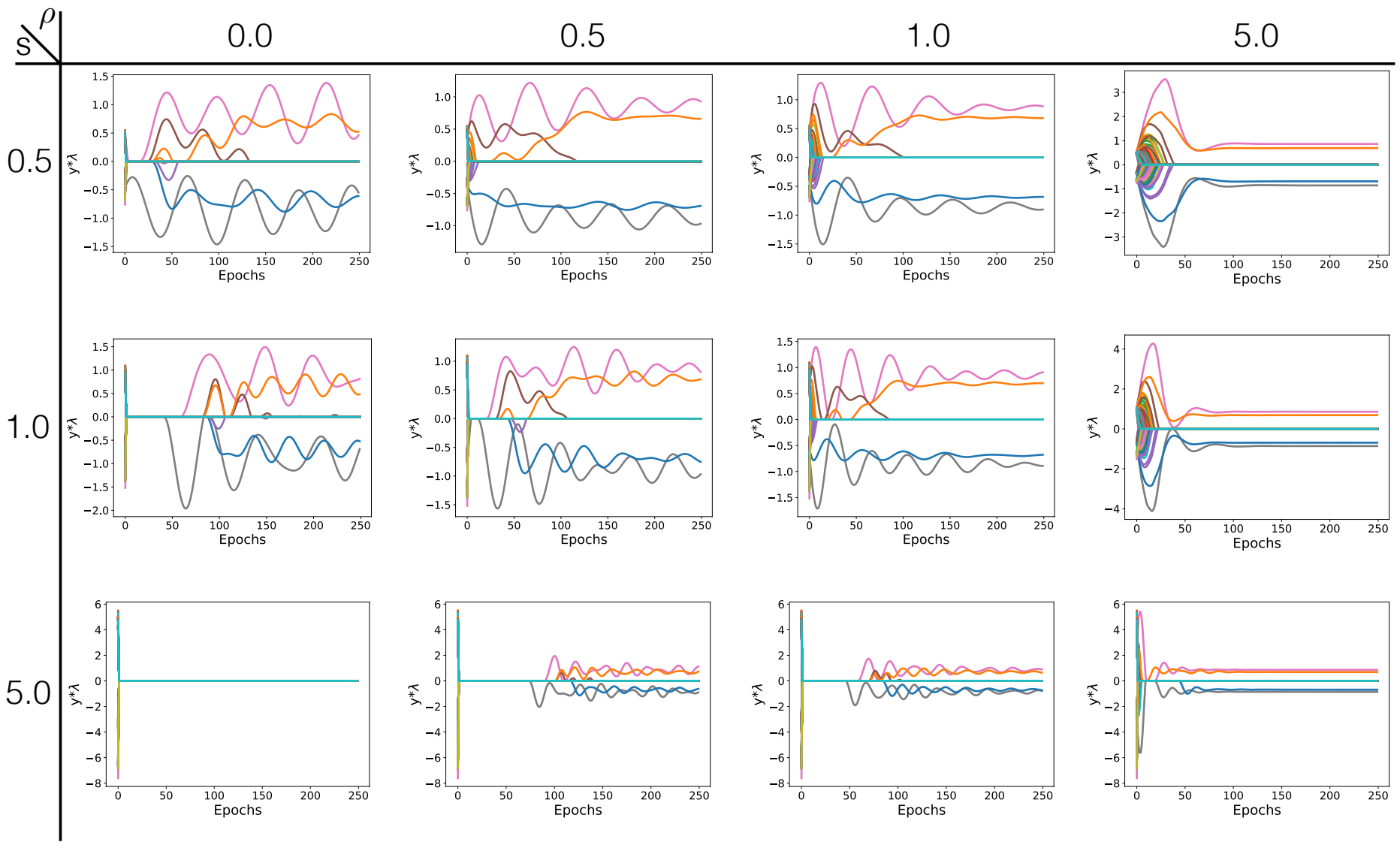

Figure 8: Here we provide an example of how $\alpha$ varies through training for $s = (0.5, 1.0, 5.0)$ and $\rho$=(0, 0.5, 1.0, 5.0) for 1 hidden layer with 100 units trained on the binary classification task described in the main text.

# D Details of the experiment for MNIST

All the experiments had constant learning rate with no momentum to avoid extraneous factors.

## D.1 Vanilla SGD:

Batch size: $Q = 128$
Learning rate: $\eta = 0.1$

## D.2 $L_2$-regularization:

Batch size: $Q = 128$
Learning rate: $\eta = 0.1$
$L_2$-coefficient grid: $\{10^{-3}, 5 \times 10^{-3}, 10^{-4}, 5 \times 10^{-4}, 10^{-5}, 5 \times 10^{-5}\}$
Final $L_2$-coefficient: $5 \times 10^{-4}$

## D.3 Minnorm, BGD:

$\rho = 0$
$s = .0002$
Learning rate: $\eta = .001$

## D.4 Minnorm, SGD:

Batch size: $Q = 128$
Augmented lagrangian co-efficient: $\rho = 0$
Lagrangian update $s = 7.8125$
Learning rate: $\eta = 10^{-5}$